\documentclass[runningheads]{llncs}

\usepackage[T1]{fontenc}

\usepackage{cite}
\usepackage{amsmath,amssymb,amsfonts}
\usepackage{algorithmic}
\usepackage{graphicx}
\usepackage{textcomp}
\usepackage{xcolor}
\usepackage{pifont}
\usepackage{subcaption}
\usepackage{hyperref}
\usepackage{multirow}
\usepackage{balance}
\usepackage{url}
\usepackage{bbding}

\hypersetup{hidelinks} 

\newcommand{\NAME}{\textsf{EFIM}}

\newcommand{\guoty}[1]{\textcolor{black}{#1}} 

\author{
Tianyu Guo\inst{1*\dag} \and Hande Dong\inst{2*}\Envelope \and Yichong Leng\inst{3} \and Feng Liu\inst{2} \and Cheater Lin\inst{2} \and Nong Xiao\inst{1} \and Xianwei Zhang\inst{1}\Envelope
}

\institute{
Sun Yat-sen University, Guangzhou, China\\
\email{guoty9@mail2.sysu.edu.cn,\{xiaon6,zhangxw79\}@mail.sysu.edu.cn} \and
Tencent, Shenzhen, China\\
\email{\{donghd66,neolscarlet\}@gmail.com,cheaterlin@tencent.com} \and
University of Science and Technology of China, Hefei, China\\
\email{lyc123go@mail.ustc.edu.cn}
}

\begin{document}

\title{EFIM: Efficient Serving of LLMs for Infilling Tasks with Improved KV Cache Reuse}

\titlerunning{EFIM: Efficient Serving of LLMs for Infilling Tasks}
\authorrunning{T. Guo et al.}

\maketitle

\begingroup\renewcommand\thefootnote{*}
\footnotetext{Equal contribution.}
\endgroup
\begingroup\renewcommand\thefootnote{\dag}
\footnotetext{This work was done during an internship at Tencent. }
\endgroup


\begin{abstract}
Large language models (LLMs) are often used for infilling tasks, which involve predicting or generating missing information in a given text. These tasks typically require multiple interactions with similar context.
To reduce the computation of repeated historical tokens, cross-request key-value (KV) cache reuse, a technique that stores and reuses intermediate computations, has become a crucial method in multi-round interactive services.  
However, in infilling tasks, the KV cache reuse is often hindered by the structure of the prompt format, which typically consists of a prefix and suffix relative to the insertion point.
Specifically, the KV cache of the prefix or suffix part is frequently invalidated as the other part (suffix or prefix) is incrementally generated.
To address the issue, we propose \NAME, a transformed prompt format of FIM to unleash the performance potential of KV cache reuse. Although the transformed prompt can solve the inefficiency, it exposes subtoken generation problems in current LLMs\guoty{, where they have difficulty generating partial words accurately.} Therefore, we introduce a fragment tokenization training method \guoty{which splits text into multiple fragments before tokenization during data processing}. Experiments on two representative LLMs show that LLM serving with \NAME~can lower the latency by 52\% and improve the throughput by 98\% while maintaining the original infilling capability. \NAME’s source code is publicly available at \url{https://github.com/gty111/EFIM}.


\end{abstract}

\setcounter{footnote}{0}

\section{Introduction}

Infilling tasks involve predicting or generating missing words, phrases, or even entire sentences within a given text. Recently, there has been a growing trend of using large language models (LLMs) like Codex \cite{HumanEval}, StarCoder \cite{starcoder,starcoder2}, CodeLlama \cite{Codellama}, Qwen2.5-coder \cite{Qwen2.5-coder} and DeepSeek-Coder \cite{deepseek-coder} for such tasks. As a result, many companies are starting to provide online services for infilling, such as OpenAI canvas \cite{canvas}, GitHub Copilot \cite{Copilot} and Amazon CodeWhisperer \cite{CodeWhisper}. However, prompts of infilling tasks require long context around the insertion point and users often need multi-turn interactions with LLMs, leading to high computational demands. Efficient serving of LLMs for infilling tasks has become an important research problem (a detailed analysis is given in \S \ref{sec:res.cost}).

\begin{figure}[t]
    \centering
    \vspace{-0.1in}
    \includegraphics[width=.8\linewidth]{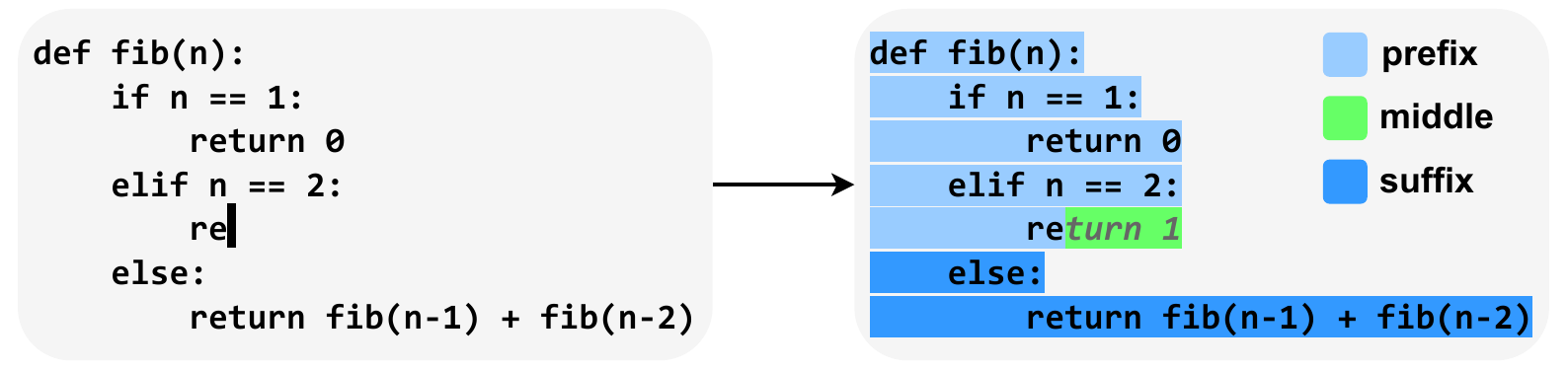}
    \caption{A python code snippet where a programmer wants to insert code inside a function. The prefix/suffix part represents content before/after the insertion point. The middle part is the content expected to infill.}
    \label{fig:code_completion_scene}
    \vspace{-0.2in}
\end{figure}

As the \textit{de facto} technique to reduce computation and accelerate LLMs inference, KV cache \cite{vLLM,DeepSpeed-Inference,Scale_Transformer} stores attention keys and values to prevent recomputation. While traditional KV cache operates within a single request, cross-request KV cache reuse\footnote{If not explicitly stated, KV cache reuse in this paper is cross-request.}\cite{sglang,CachedAttention,ChunkAttention,Pensieve} has been proposed to minimize redundant KV cache recomputation in multi-turn services \cite{MINT}, significantly reducing latency. However, cross-request KV cache reuse imposes strict constrains that prefix of prompt tokens must remain identical. In the infilling scenario as shown in Figure \ref{fig:code_completion_scene}, the prompt typically follows the fill-in-the-middle (FIM) format \cite{FIM}, i.e., ``$<$P$>$prefix$<$S$>$suffix$<$M$>$'', where $<$P$>$, $<$S$>$ and $<$M$>$ are FIM special tokens to connect prefix, suffix and middle parts. A common behavior in infilling tasks is the continuous expansion of the prefix, which invalidates the KV cache of the suffix. This occurs because incremental changes in the prefix alter the preceding tokens of the suffix. \guoty{Thus, KV cache of the suffix need to be recomputed in each interaction, requiring high computation resources.} To solve it, we propose transforming the FIM format from ``$<$P$>$prefix+$inc$$<$S$>$suffix$<$M$>$'' to ``$<$P$>$prefix$<$S$>$suffix$<$M$>$$inc$'' (\NAME), where $inc$ represents the incremental prefix change. This modification ensures that both the prefix and suffix remain unchanged, with the variation confined to $inc$. Consequently, KV cache reuse can be extended from solely the prefix to include both the prefix and suffix.

\guoty{Despite \NAME~improves KV cache reuse, it reveals a hidden subtoken\footnote{In the paper, we refer to incomplete words as subtokens like ``pri'' in ``print''. Incomplete words caused by tokenizer are not included.} generation problem in current LLMs. The issue stems from \NAME's requirement for models to generate subtokens after $inc$, a capability not supported by existing LLMs.
To enable universal subtoken generation, we propose a fragment tokenization training method, involving randomly splitting sentences into multiple segments, tokenizing each segment individually, and then concatenating the results. In this way, the model can learn the ability to generate the remaining subtokens based on the initial subtoken during training, thereby addressing the subtoken issue encountered by \NAME.}


In summary, the contributions of this paper are:
\begin{itemize}
    \item We identify that the efficiency of LLM inference for infilling tasks is hindered by the FIM format, as the KV cache of the prefix/suffix part is frequently invalidated by the growing suffix/prefix.
    \item We propose \NAME, the first method to transform the FIM prompt format, unlocking the potential of KV cache reuse.
    \item To enhance subtoken generation ability, we introduce a fragment tokenization training method on data processing.
    \item Experiments on two pretrained LLMs show that EFIM reduces average latency by 52\% and increases throughput by 98\%, while preserving model capability.
\end{itemize}
\section{BACKGROUND AND MOTIVATION}

\subsection{Training LLMs with FIM}
\begin{figure}[h]
    \centering
    \vspace{-0.3in}
    \includegraphics[width=.7\linewidth]{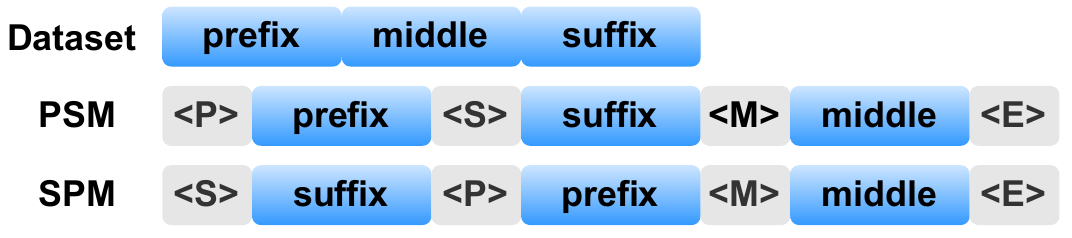}
    \caption{Comparison of PSM and SPM. $<$P$>$ follows prefix part, $<$S$>$ follows suffix part, $<$M$>$ follows middle part and $<$E$>$ marks the end of infilling span.}
    \label{fig:psm_spm}
    \vspace{-0.2in}
\end{figure}

Current decoder-based autoregressive (AR) language models \cite{GPT3,Bert,PaLM,Gemini} are capable of generating text from left to right. However, they struggle with infilling tasks, where the model is required to generate text at a specific location within a snippet, conditioned on both a prefix and a suffix. To address this limitation, FIM capabilities have been integrated into AR models without compromising their standard left-to-right generation \cite{FIM,InCoder,CodeGen,CodeGen2}. The core idea of FIM involves splitting the documents into three parts, and then relocating the middle part to the end. Models are trained on a mixture of FIM transformed data and standard left-to-right data. As shown in Figure \ref{fig:psm_spm}, FIM can be prepared in two ways denoted as prefix-suffix-middle (PSM) and suffix-prefix-middle (SPM). In general, the LLMs can own both abilities.

\subsection{KV Cache Reuse Inefficiency with FIM}

\begin{figure}[h]
    \centering
    \includegraphics[width=.9\linewidth]{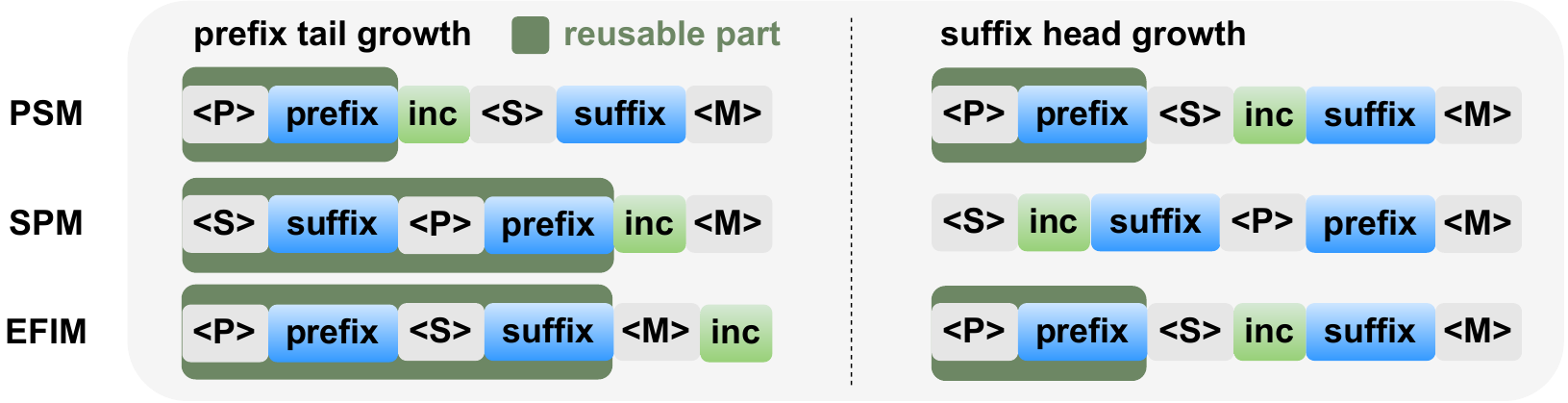}
    \caption{Reusable part between PSM, SPM and \NAME~when the growth ($inc$) happens either at the prefix tail or at the suffix head. The reusable part includes the content before $inc$.}
    \label{fig:reuse_comp}
    \vspace{-0.2in}
\end{figure}

LLMs notably feature their self-attention mechanism, and the KV cache is used to accelerate the inference \cite{Transformer,flashAttention,flashAttention2,InfiniGen,Tabi,Orca,AttentionSink,H2O}. Additionally, the KV cache of shared prefix across different sequences can be reused to avoid redundant computations \cite{CachedAttention,PromptCache,ChunkAttention,Pensieve}. In infilling scenarios, users usually need to engage in multi-round interactions with LLMs, especially when dealing with long contexts. According to our statistics from online infilling services, most of modifying behaviors involve appending tokens to the tail of the prefix or the head of the suffix. Figure \ref{fig:reuse_comp} illustrates the reusable parts of the KV cache across different prompt formats. It shows that changes to the tail of the prefix invalidate the KV cache of the suffix in PSM, while changes to the head of the suffix invalidate both the prefix and suffix in SPM. To address unnecessary KV cache invalidation, we propose \NAME, which relocates the prefix increment to the end of the prompt in PSM. \guoty{\NAME~combines the advantage of PSM and SPM, achieving the most KV cache reuse in both scenes (prefix tail growth and suffix head growth).}


\begin{figure}[ht]
    \centering
    \vspace{-0.2in}
    \begin{subfigure}[b]{.48\linewidth}
        \centering
        \includegraphics[width=\linewidth]{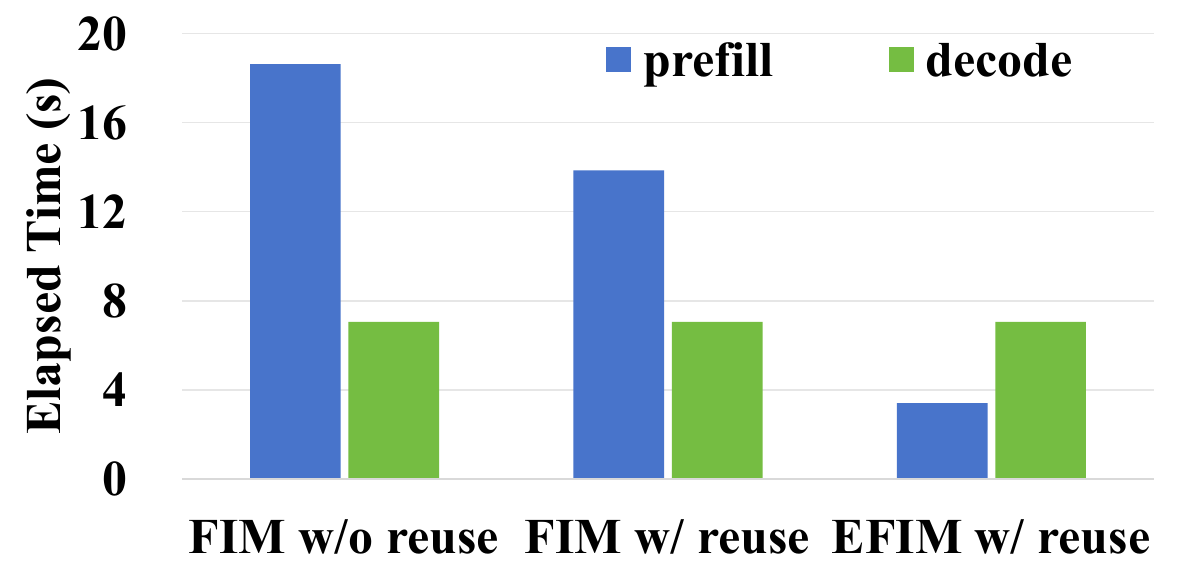}
        \caption{Deepseek-coder-6.7B}
    \end{subfigure}
    \begin{subfigure}[b]{.48\linewidth}
        \centering
        \includegraphics[width=\linewidth]{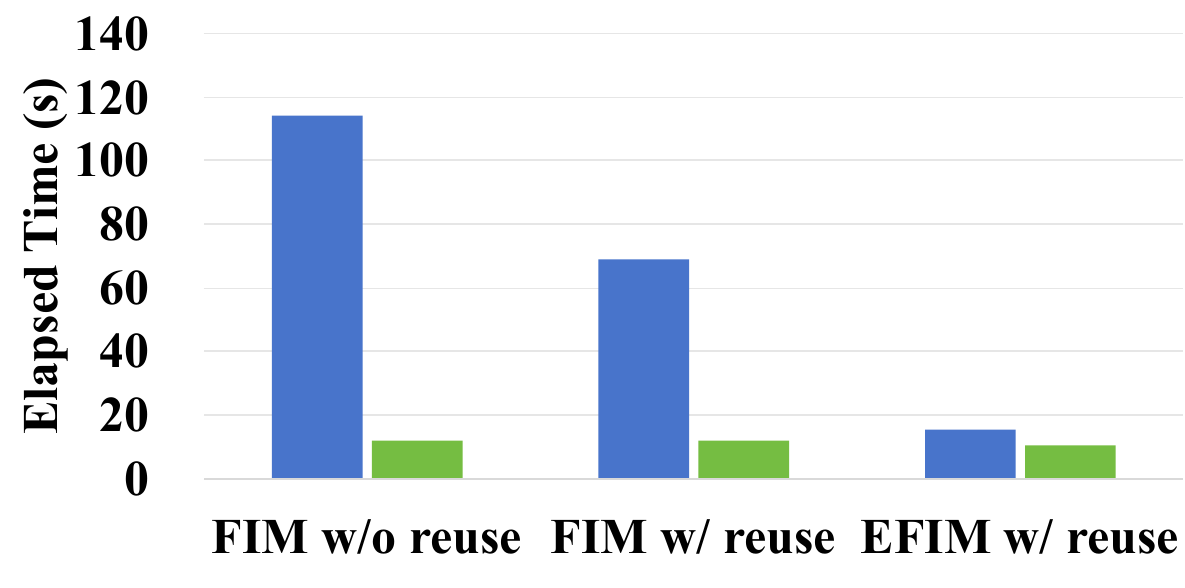}
        \caption{Llama3.1-8B}
    \end{subfigure}
    \caption{Elapsed time breakdown of prefill and decode stage for infilling serving (average input/output length is 2100/32) between FIM and \NAME.}
    \label{fig:prefill_decode}
    \vspace{-0.2in}
\end{figure}

We also analyze the end to end elapsed time breakdown for infilling serving as illustrated in Figure \ref{fig:prefill_decode}. Without cross-request KV cache reuse (FIM w/o reuse), the prefill overhead can be up to 9 times (114 vs. 12) than decode. Even with KV cache reuse (FIM w/ reuse), the time gap between prefill and decode remains significant, reaching up to 6 times (69 vs. 12). With \NAME, the overhead of prefill stage is significantly reduced by 40\% on average (114 vs. 69) compared to FIM w/ reuse. This demonstrates \NAME's superior computational efficiency during the prefill phase.

\subsection{Subtoken Generation Capability with LLMs}
\label{sec:mot_subword}
\begin{figure}[h]
    \centering
    \vspace{-0.1in}
    \includegraphics[width=.8\linewidth]{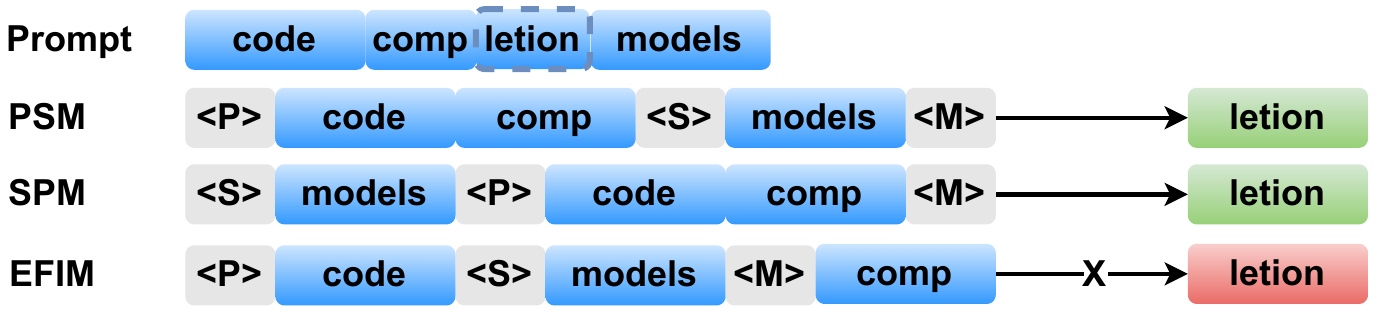}
    \caption{Subtoken generation ability between different prompt formats considering prompt ``code comp[] models''.}
    \label{fig:subword}
    \vspace{-0.2in}
\end{figure}

During the pre-training of LLMs, the models are typically trained on vast corpora of text data to assimilate the statistical regularities and semantic representations of language. Despite their prowess in generating coherent text, LLMs exhibit limitations when it comes to handling subtoken generation tasks due to the lack of relevant cases in their training data. \guoty{For instance, existing models fail to generate ``nt'' after ``pri''. Whereas} infilling LLMs overcome this limitation by training on documents split into three parts and joined with FIM special tokens.
This process introduces subtokens into the dataset, as the splits created by the FIM special tokens often result in partial tokens (subtokens). Therefore, subtokens typically appear around FIM special tokens, i.e., ``subtoken<M>subtoken'', and their generation relies on the context provided by these tokens. Without this context, the model loses the ability to generate subtokens effectively. For example, as shown in Figure 5, when the input prompt is ‘code comp[] models’, where ‘[]’ represents the missing content, both PSM and SPM can successfully generate the subtoken ‘letion’. However, \NAME~fails to generate sutokens correctly when the prefix ends with a subtoken, highlighting the limitations of directly applying LLMs in such cases. To address this challenge, we must enhance LLMs with a universal subtoken generation capability, ensuring that the model can generate subtokens regardless of the presence of FIM special tokens.


\section{DESIGN}

\begin{figure}[ht]
    \centering
    \vspace{-0.2in}
    \includegraphics[width=.9\linewidth]{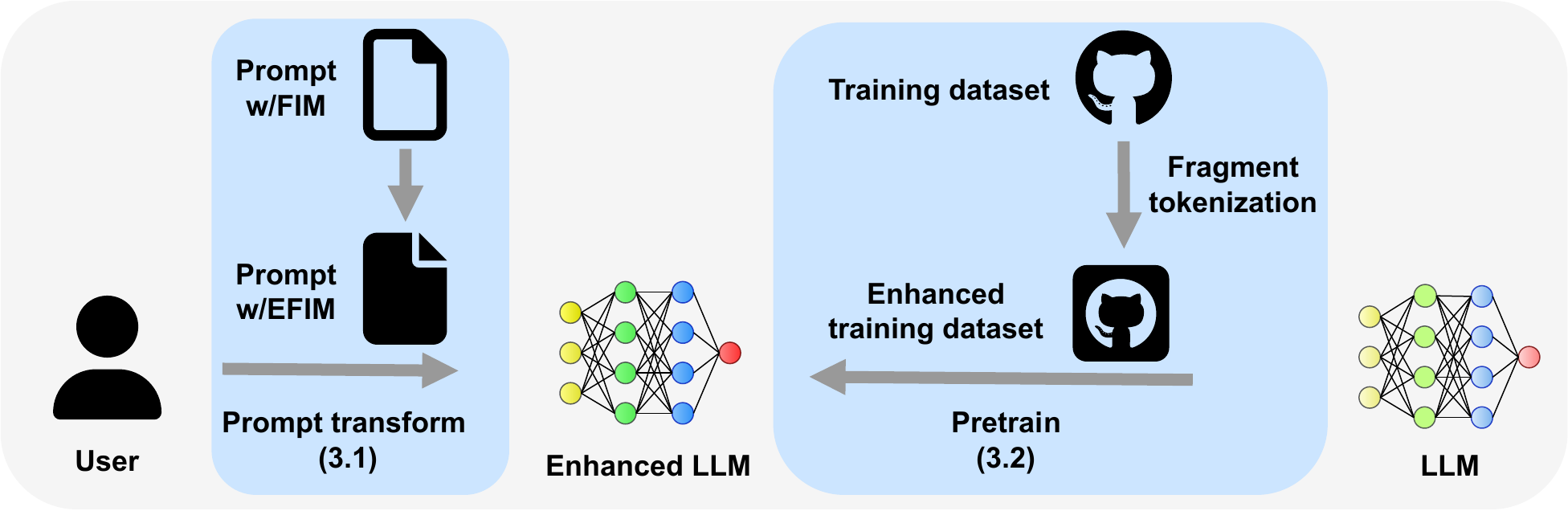}
    \caption{Overall diagram of design with \NAME.}
    \label{fig:design_overall}
    \vspace{-0.1in}
\end{figure}

Our proposed design with \NAME~consists of two key parts as illustrated in Figure \ref{fig:design_overall}. The first part operates between the user and the LLM to seamlessly and automatically convert the prompt format from FIM to \NAME. This transformation is fully transparent to the user, ensuring a smooth and intuitive experience.
The second part introduces a fragment tokenization training method focused on data processing. This method is designed to augment the LLM’s ability to generate subtokens, a critical requirement for \NAME~functionality.
Our implementation introduces no architectural changes, making \NAME~accessible for integration into existing LLM frameworks.

\vspace{-0.1in}
\subsection{From FIM to \NAME}
\label{sec:design_FIMX}

\begin{figure}[h]
    \centering
    \vspace{-0.3in}
    \includegraphics[width=.7\linewidth]{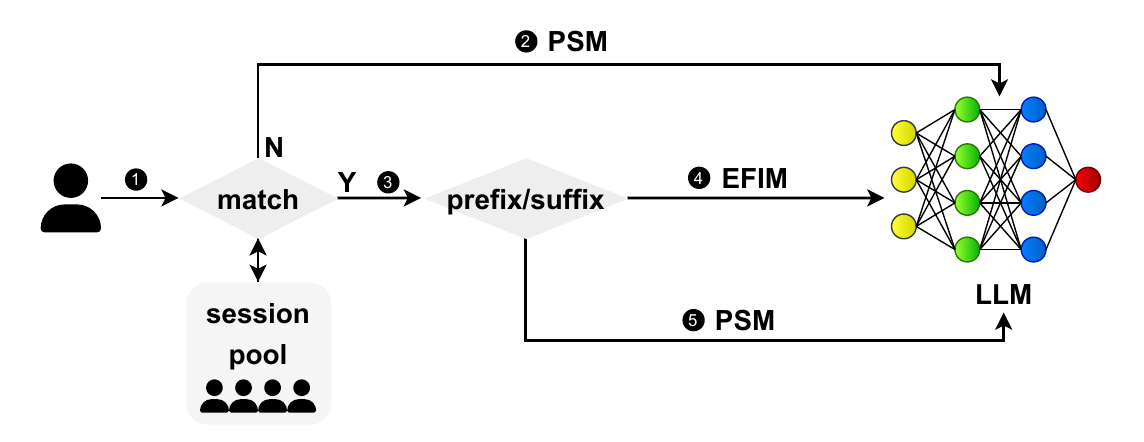}
    \caption{The prompt transformation process from FIM to \NAME.}
    \label{fig:FIM_to_EFIM}
    \vspace{-0.2in}
\end{figure}

To automatically convert prompt format from FIM to EFIM, we use a per-user session pool to track the most recent interaction between users and the LLMs as shown in Figure \ref{fig:FIM_to_EFIM}. Each session stores the prefix and suffix parts extracted from the user's previous request. 
\ding{182} When a new request is received, we first check if the user has an existing session in the pool and identify the prefix and suffix parts. \ding{183} If no matching session is found, we forward the prompt in PSM format to the LLM inference engine and create a new session for the user. 
\ding{184} If a matching session is located, we compare the prefix/suffix in the new request with the one from the previous interaction. 
\ding{185} If the prefix in the new request contains additional content compared to the session prefix, we split the new prefix into a common part and an incremental part referred to as $inc$. We then construct the EFIM-formatted prompt by concatenating the common part, the new suffix, and $inc$, before sending it to the LLM. In this way, the incremental prefix content does not invalidate the KV cache for the suffix, unlike in the PSM format.
\ding{186} If the suffix of new request has an incremental part compared to the session suffix, we send the request in PSM format directly to the LLM inference engine. In this scenario, the KV cache for the common prefix can still be reused, offering an advantage over the SPM format.

\subsection{Fragment Tokenization Training Method}
\label{sec:design_train}

\begin{figure}[h]
    \centering
    \vspace{-0.1in}
    \includegraphics[width=.7\linewidth]{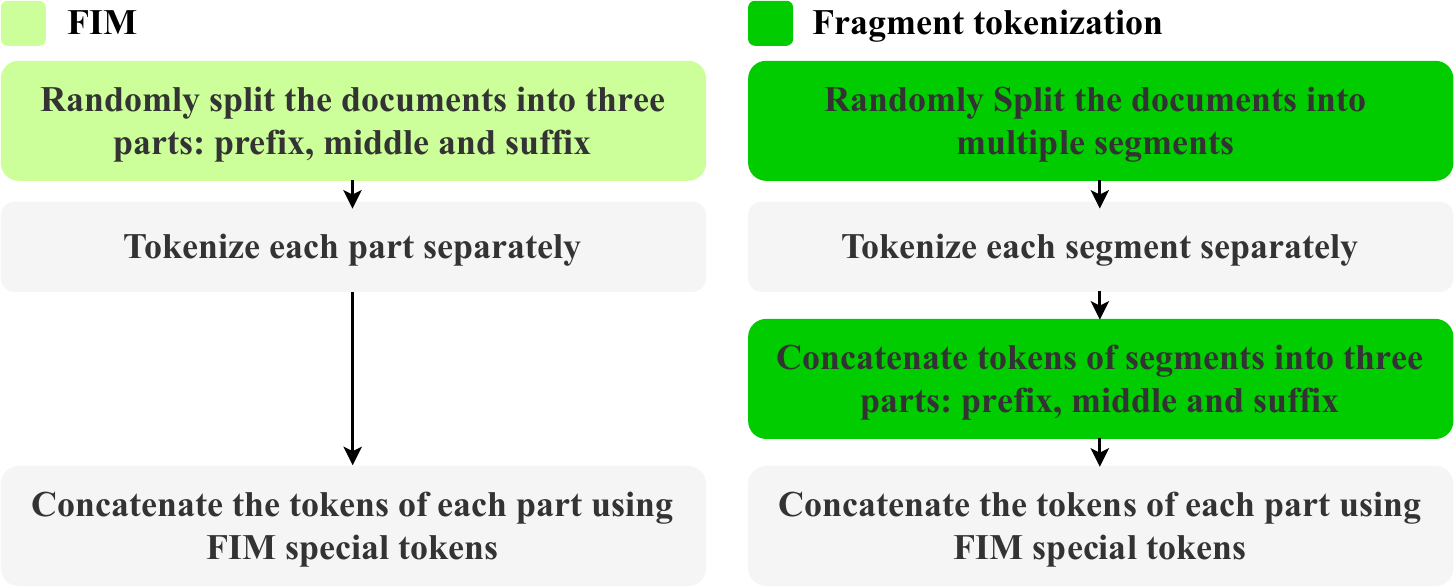}
    \caption{Data processing diagram between FIM (left) and fragment tokenization (right). The length of each segment follows uniform distribution [1,200].}
    \label{fig:data_process_diagram}
    \vspace{-0.1in}
\end{figure}

\begin{figure}[h]
    \centering
    \includegraphics[width=.8\linewidth]{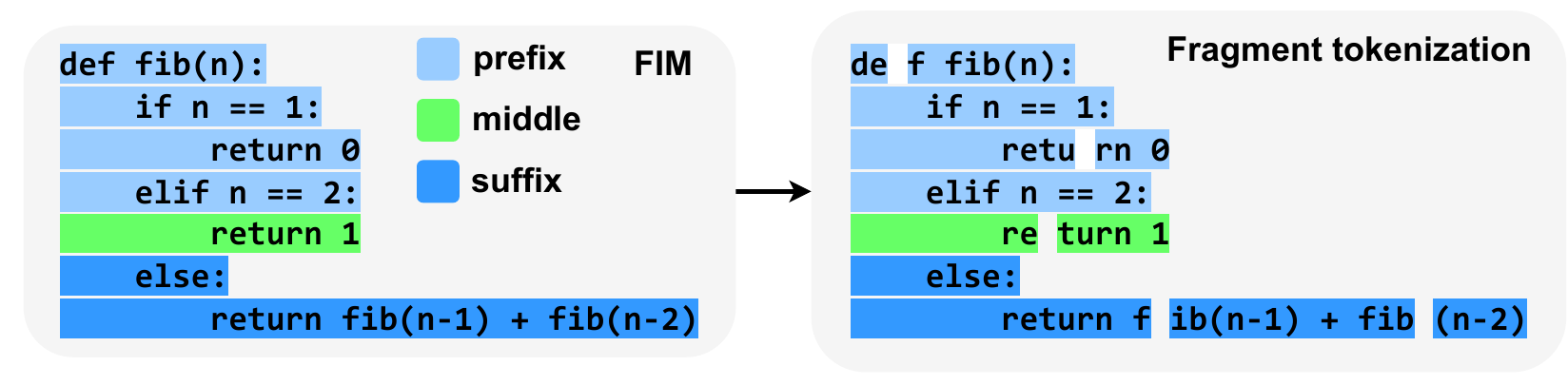}
    \caption{Comparison between FIM (left) and fragment tokenization (right) data processing examples. FIM-based method only splits the text into three parts, while our method splits the text into multiple segments and employ tokenization for each segment.}
    \label{fig:comp_data_process}
    \vspace{-0.2in}
\end{figure}

To equip LLMs with universal subtoken generation capability, we propose a novel fragment tokenization training method focused on data processing. It fundamentally differs from FIM in how the training dataset is processed. Figure \ref{fig:data_process_diagram} shows the similarities and differences parts between the two approaches. Both FIM and our method apply the transformation to the documents to adjust the order of prefix, suffix and middle. While FIM directly tokenize the three parts, our method split the text into multiple segments to allow subtokens to be generated at more locations. We also provide an example on data processing in Figure \ref{fig:comp_data_process}.



The fragment tokenization approach allows subtokens to appear not only adjacent to FIM special tokens but also throughout any position in the sequence. As a result, the model develops a more comprehensive and universal subtoken generation capability. By embedding subtokens across varied contexts within the training data, the enhanced LLM becomes better equipped to generate subtokens seamlessly in diverse scenarios, making it far more versatile and effective for real-world applications. \emph{It is important to note that our approach can serve as a drop-in replacement of current LLM training process, incurring no additional overhead. For existing LLMs, our method can be applied during continued pretraining.}

\section{EXPERIMENTAL METHODOLOGY}

We conduct continue pretraining with 64 A100 GPUs on two representative LLMs, Deepseek-coder-6.7B\footnote{This enhanced model has been used in production for AI Code Assistant.} \cite{deepseek-coder} and Llama3.1-8B \cite{llama3}, using fragment tokenization method to enhance their sub-token generation ability. 
The pretraining process for each model takes less than a week. Notably, the additional overhead can be avoided if the fragment tokenization training method is applied from the beginning.
The training dataset consists of 108 billion tokens collected from StarCoderData \cite{starcoderdata}. 
For Llama3.1-8B, we pretrain a baseline version (based on the original LLM) to equip it with FIM ability. \guoty{The experiments mainly focus on three questions: 
\begin{enumerate}
    \item Does fragment tokenization method impact infilling ability and truely make LLMs possess subtoken generation ability? (\S \ref{sec:expr_code_compl} and \S \ref{sec:res_infill_subtoken})
    \item Can \NAME~improve the KV cache reuse and the efficiency of LLM serving? (\S \ref{sec:expr_speedup} and \S \ref{sec:res_speedup})
    \item Is it worth the training overhead to gain inference speed? (\S \ref{sec:res.cost})
\end{enumerate}
}

\subsection{Infilling and Subtoken Generation Ability}
\label{sec:expr_code_compl}
\begin{figure}[ht]
    \centering
    \vspace{-0.2in}
    \includegraphics[width=\linewidth]{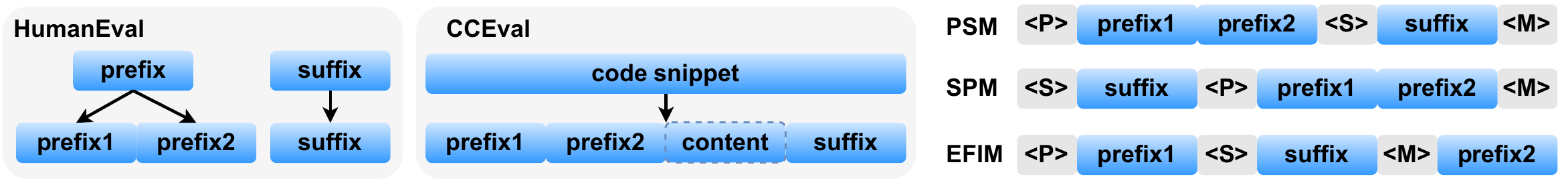}
    \caption{Prompt creation procedure and prompt format between FIM and \NAME.}
    \label{fig:benchmark_prompt}
    \vspace{-0.2in}
\end{figure}

Current infilling evaluations rely on FIM, which is incompatible with \NAME. To assess both infilling and universal subtoken abilities with \NAME, we adapt the prompt format from HumanEval Infilling \cite{FIM} and CrossCodeEval (CCEval) \cite{CCeval}, focusing on the scenario where tokens are appended to the prefix. This scenario highlights the behavioral differences between PSM and \NAME.


\textbf{Prompt creation.}
Figure \ref{fig:benchmark_prompt} illustrates the prompt creation process and prompt format of FIM and \NAME.
Based on HumanEval Infilling, we randomly split the prefix into prefix1 and prefix2, use prefix1 as the original prefix and prefix2 as the increment of prefix. Note that, HumanEval Infilling includes three infiling benchmarks, single-line, multi-line and random-span. In the single-line and multi-line benchmarks, prefix2 does not end with subtokens because they require the generation of complete single or multiple lines of code. In contrast, the random-span benchmark may have subtokens at the end of prefix2.
In CCEval, we modify the prompt format by randomly splitting the entire code snippet into four parts (prefix1, prefix2, content to infill and suffix). Prefix1 is used as the original prefix, suffix as the original suffix and prefix2 as the increment of prefix. Since the splitting process is entirely random, prefix2 may end with subtokens. 

\subsubsection{Metrics}
We use pass@1 for HumanEval Infilling, EM and ES for CCEval.
\begin{itemize}
    \item \textbf{Pass@1}: One code sample is generated per problem, a problem is considered solved if the sample passes the unit tests, and the percent of problems solved is reported.
    \item \textbf{Exact Match (EM)}: The percent of the situations when the generated code is exact the ground truth.
    \item \textbf{Edit Similarity (ES)}: Similarity score between the generated code and the ground truth using the Levenshtein distance algorithm. The score ranges from 0 to 100, where higher values indicate greater similarity. 
\end{itemize}

\subsubsection{Schemes}
We conduct a comparative analysis between the original LLM (oLLM) and our proposed enhanced LLM (eLLM), both of which utilize FIM or \NAME.

\subsection{Inference Speedup}
\label{sec:expr_speedup}
To evaluate the efficiency of different reusable KV cache levels (none, prefix, and prefix+suffix), we compare \NAME~with FIM\footnote{The advantage of \NAME~compared to FIM can be seen different reusable KV cache levels. Therefore, in this experiment, we focus on the efficiency of serving at different reusable KV cache levels.} in a scenario where tokens are appended to the prefix. This setup simulates infilling cloud services, where multiple users interact with LLMs over several rounds. In each round, a prefix is extended with new tokens.
Instead of a fixed request rate, an unrealistic scenario, we adjust the service load based on the number of users. Each user acts as an individual client, sending a request for the next round only after receiving the previous response. For our experiment, we set the number of rounds to 5 and the number of users to 16. The average input/output length is 2355/128.

\subsubsection{Environment}
We utilize the vLLM inference framework (v0.6.2) \cite{vLLM}.
The experiments are performed on a server with an AMD EPYC 7742 processor, 256GB of host memory and an NVIDIA A100 GPU.

\subsubsection{Metrics}
\begin{itemize}
    \item \textbf{Latency}: Average end to end latency for each request.
    \item \textbf{Input throughput}: Average input token processing throughput.
    \item \textbf{Request throughput}: Average request completion rate.
    \item \textbf{Reuse rate}: Cross-request KV cache reuse rate.
\end{itemize}

\subsubsection{Schemes}
\begin{itemize}
    \item \textbf{Baseline}: PSM without KV cache reuse.
    \item \textbf{FIM}: PSM with KV cache reuse.
    \item \textbf{\NAME}: \NAME~with KV cache reuse.
\end{itemize}
\section{RESULTS AND ANALYSIS}
\subsection{Infilling and Subtoken Generation Ability}
\label{sec:res_infill_subtoken}


\begin{table}[ht]
\setlength{\tabcolsep}{3pt}
\renewcommand\arraystretch{1.5}
    \centering
    \vspace{-0.1in}
    \caption{Evaluation results on infilling benchmarks. The left part shows Pass@1 rate (higher is better) in HumanEval Infilling where S stands for single-line, M stands for multi-line and R stands for random-span. The right part shows EM and ES metric (higher is better) in CCEval. oLLM and eLLM abbreviate for LLM training with FIM and fragment tokenization, respectively. The underlined numbers indicate a decrease in the model's ability due to the lack of subtoken generation capability.}
    \vspace{0.1in}
    \begin{tabular}{c|cccccc|cccc}
        \hline
        \textbf{Benchmark} & \multicolumn{6}{c|}{\textbf{HumanEval Infilling}} & \multicolumn{4}{c}{\textbf{CCEval}} \\
       \textbf{Model} & \multicolumn{3}{c}{\textbf{Deepseek}} & \multicolumn{3}{c|}{\textbf{Llama}} & \multicolumn{2}{c}{\textbf{Deepseek}} & \multicolumn{2}{c}{\textbf{Llama}} \\
         & S & M & R & S & M & R & EM & ES & EM & ES \\
        \hline
        \hline
        oLLM w/FIM & 89.64 & 61.96 & 76.77 & 87.32 & 56.90 & 62.99 & 33.51 & 78.43 & 29.40 & 71.30 \\
        oLLM w/EFIM & 90.03 & 62.25 & \underline{52.44} & 86.35 & 56.54 & \underline{38.35} & \underline{11.19} & \underline{71.04} & \underline{6.82} & \underline{53.44} \\
        eLLM w/FIM & 88.48 & 61.62 & 75.12 & 87.12 & 57.73 & 67.20 & 33.27 & 79.24 & 31.51 & 71.15 \\
        eLLM w/EFIM & 89.64 & 62.82 & 75.61 & 86.83 & 56.35 & 64.27 & 32.51 & 78.91 & 30.91 & 70.48 \\
         \hline
    \end{tabular}
    \vspace{-0.1in}
    \label{tab:humaneval_cceval}

\end{table}



Table \ref{tab:humaneval_cceval} presents the evaluation results of infilling performance. 
For the HumanEval Infilling single/multi-line benchmark, the pass rates between \textit{oLLM w/FIM} and \textit{oLLM w/\NAME}~remain close (with a difference of less than 1\%) as the single/multi-line tasks do not require subtoken generation (there are no subtokens at the end of the prefix increment). This demonstrates that \NAME~has little influence when subtoken generation is not required. However, for the random-span benchmark, the pass rate drops significantly by 24\% from \textit{oLLM w/FIM} to \textit{oLLM w/\NAME}, highlighting the model's inability to generate subtokens. In contrast, \textit{eLLM w/\NAME}~can maintain equivalent performance compared to \textit{oLLM w/FIM}, indicating that the fragment tokenization method (\S \ref{sec:design_train}) can effectively solve subtokens generation problems. \textit{eLLM w/FIM} also exhibits close performance compared to \textit{oLLM w/FIM}, showing that the fragment tokenization method has little impact on infilling ability.
For CCEval, the metrics shows similar pattern. Compared to ES, EM shows a more significant decrease as the model struggles to generate subtokens but performs well in generating other types of content.

\subsection{Inference Speedup}
\label{sec:res_speedup}

\begin{figure}[h]
    \vspace{-0.1in}
    \centering
    \begin{subfigure}[b]{.42\linewidth}
        \centering
        \includegraphics[width=\linewidth]{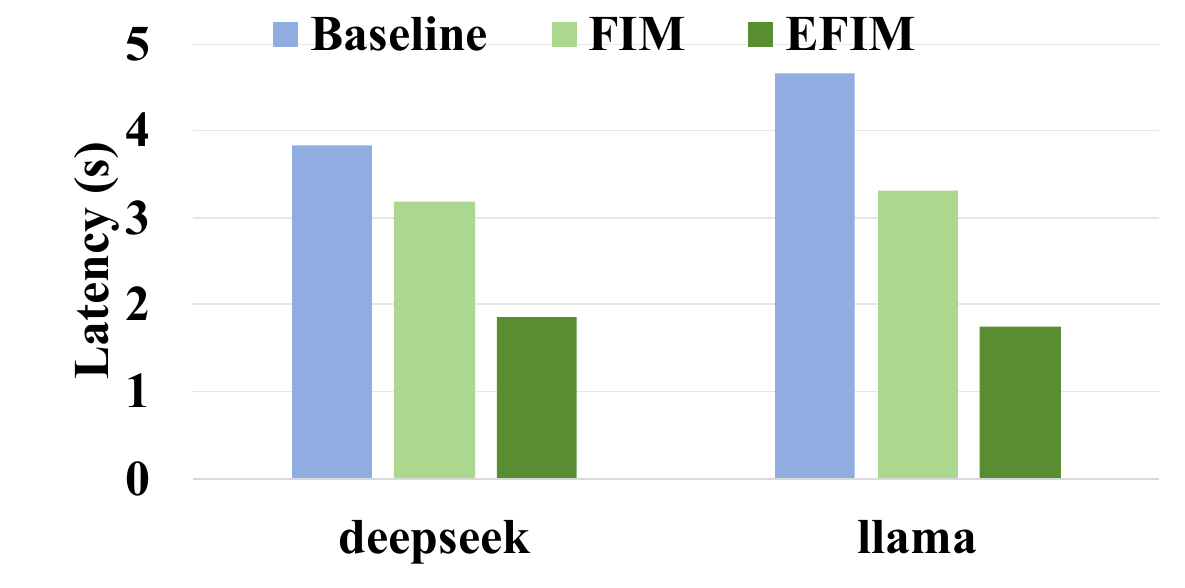}
        \caption{Average latency}
    \end{subfigure}
    \begin{subfigure}[b]{.42\linewidth}
        \centering
        \includegraphics[width=\linewidth]{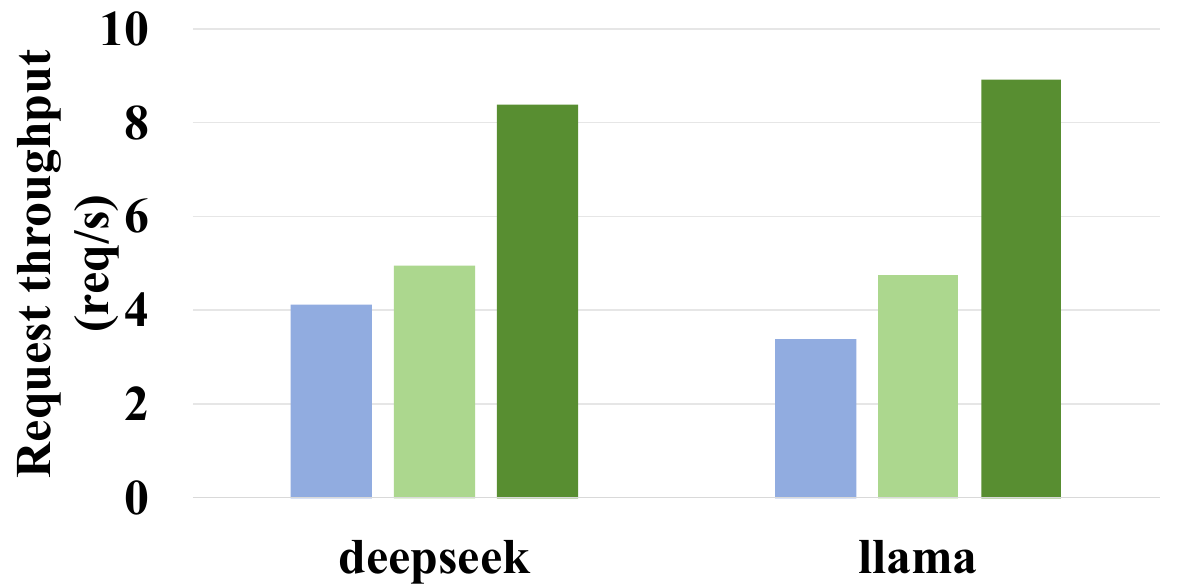}
        \caption{Request throughput}
    \end{subfigure}
    \begin{subfigure}[b]{.42\linewidth}
        \centering
        \includegraphics[width=\linewidth]{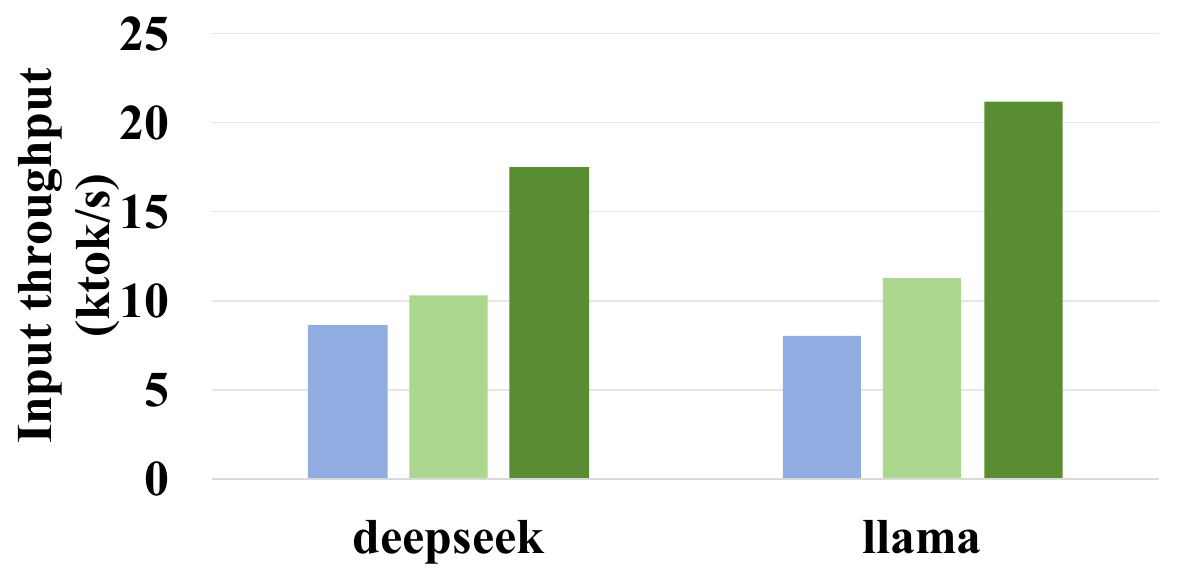}
        \caption{Input token throughput}
    \end{subfigure}
    \begin{subfigure}[b]{.42\linewidth}
        \centering
        \includegraphics[width=\linewidth]{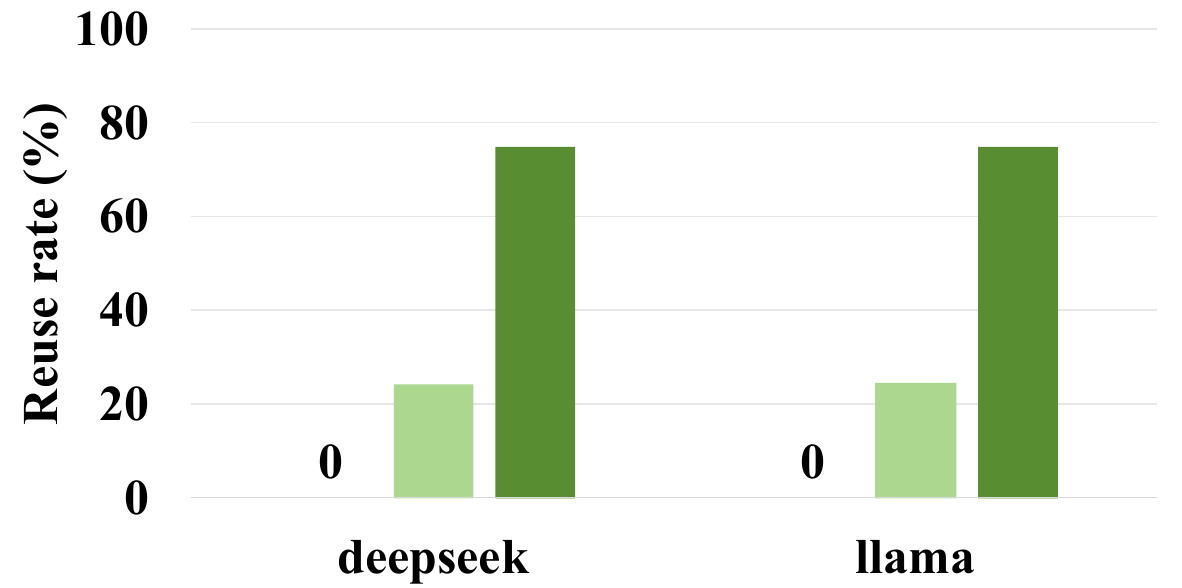}
        \caption{KV cache reuse rate}
    \end{subfigure}
    \caption{Overall inference performance on average latency, request throughput, input token throughput and KV cache reuse rate to illustrate the efficiency of different degrees of reusable KV cache.}
    \label{fig:overall_metric}
    \vspace{-0.2in}
\end{figure}


Figure \ref{fig:overall_metric} illustrates the overall inference performance. Among the three schemes, \textit{Baseline} performs the worst due to the lack of KV cache reuse, requiring the entire prompt's KV cache to be recomputed in each round which is highly time consuming. Instead, \textit{FIM} reduces latency by 21\% and improves throughput by 26\% on average by avoiding the recomputation of the prefix's KV cache. However, it still requires recomputing the suffix's KV cache due to the inefficiency of FIM. \textit{\NAME}~addresses this issue, achieving an average latency reduction of 52\% and a throughput increase of 98\%. Besides, the average latency per request drops below 2 seconds, significantly enhancing user experience. 
\textit{\NAME}~achieves the lowest latency and highest throughput by maximizing KV cache reuse, as evidenced by the highest input token throughput.

\begin{figure}[t]
    \vspace{-0.1in}
    \centering
    \begin{subfigure}[b]{.42\linewidth}
        \centering
        \includegraphics[width=\linewidth]{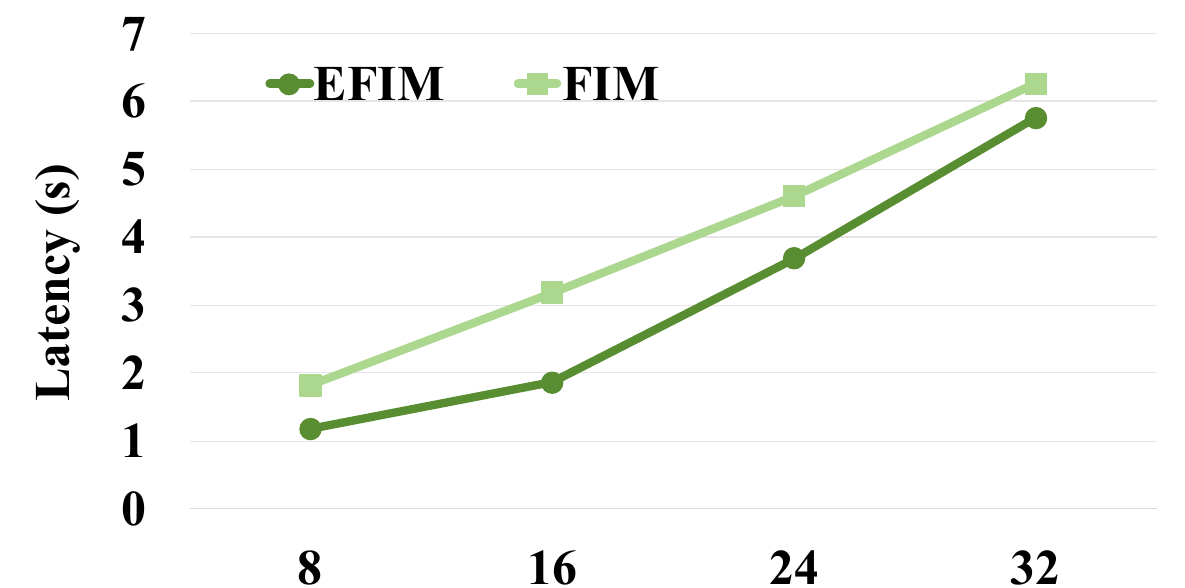}
    \end{subfigure}
    \begin{subfigure}[b]{.42\linewidth}
        \centering
        \includegraphics[width=\linewidth]{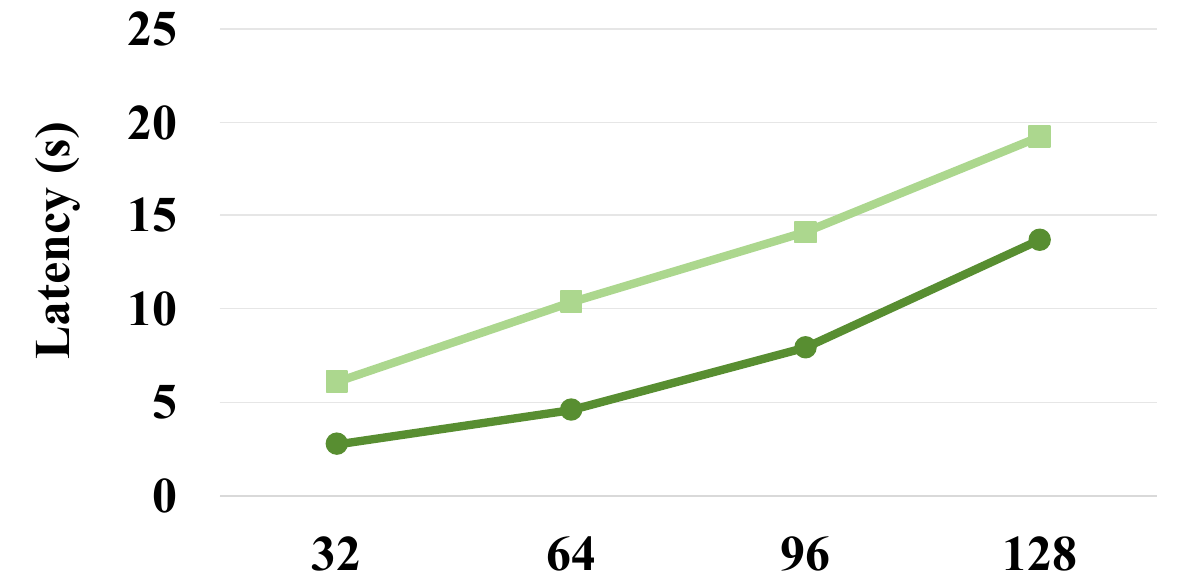}
    \end{subfigure}
    
    \begin{subfigure}[b]{.42\linewidth}
        \centering
        \includegraphics[width=\linewidth]{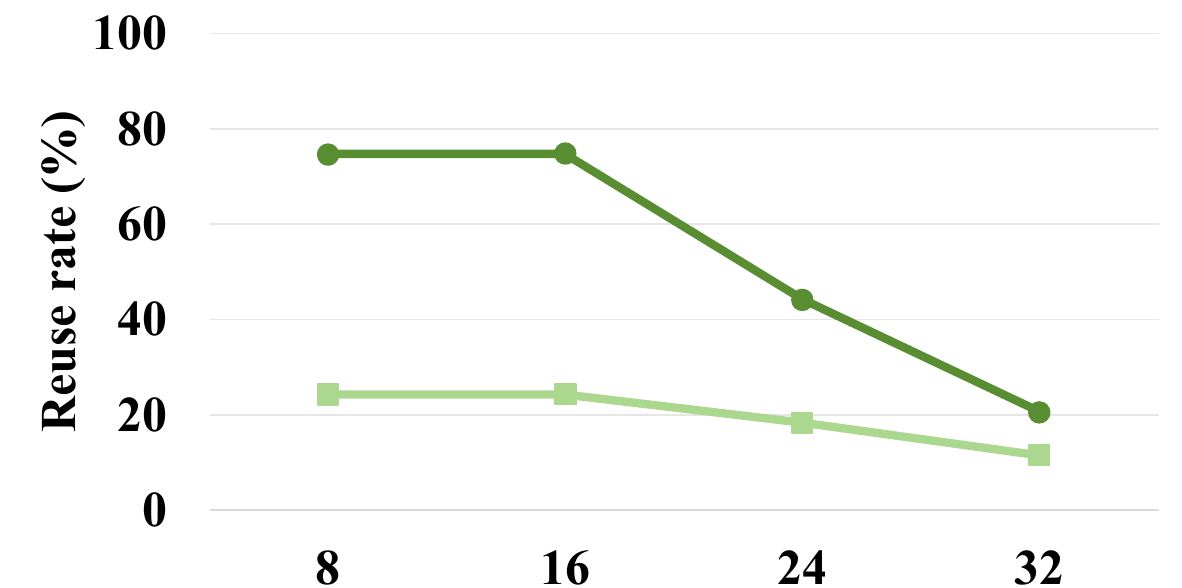}
        \caption{deepseek}
    \end{subfigure}
    \begin{subfigure}[b]{.42\linewidth}
        \centering
        \includegraphics[width=\linewidth]{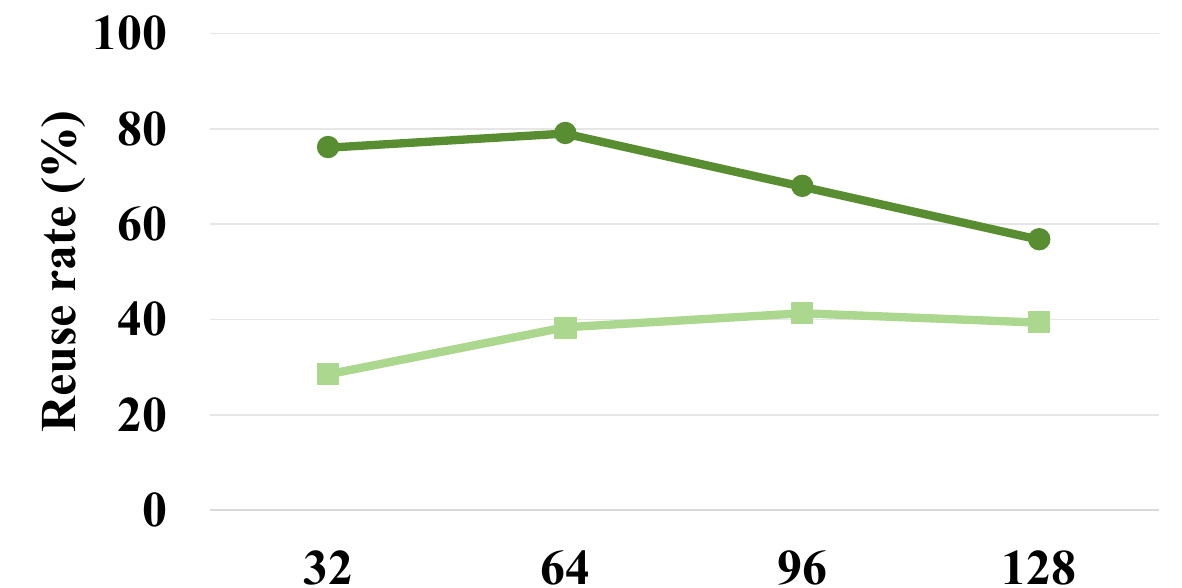}
        \caption{llama}
    \end{subfigure}
    \caption{Variation of latency (above) and KV cache reuse rate (below) as the number of users (horizontal axis) increases.}
    \label{fig:user_sensitivity}
    \vspace{-0.2in}
\end{figure}

\textbf{Number of concurrently serving users.} To evaluate the impact of the number of concurrently serving users, we conduct a sensitivity study. Figure \ref{fig:user_sensitivity} illustrates the average latency and KV cache reuse rate as user count increases. From the results, we observe that the latency of \textit{FIM} increases almost proportionally with the number of users. In contrast, \textit{\NAME}~exhibits a steeper latency curve as the user count grows, which can be attributed to a significant decline in its KV cache reuse rate. When the number of users is relatively low, \textit{\NAME}~maintains a stable reuse rate of around 80\%. However, as the user count increases, the total capacity of the KV cache for completed requests gradually exceeds available GPU memory, leading to a drop in reuse rate. On the other hand, \textit{FIM} consistently shows a lower reuse rate, remaining below 40\% across all user counts.


\subsection{Cost Efficiency}
\label{sec:res.cost}

While existing LLMs require continued pretraining to enable subtoken generation abilities, our method demonstrates superior cost efficiency. For instance, Meta's Llama3.1-8B model requires 1.46 million H100 GPU hours for training \cite{LlamaModelCard}. In contrast, our fragment tokenization approach consumes only 10,752 A100 GPU hours ($64\times7\times24$), representing merely 0.74\% of training cost of Llama3.1-8B. According to the Deepseek technical report \cite{DeepSeekV3,DeepSeekV3Serving}, the Deepseek V3 model requires 2.788 million H800 GPU hours for training, with an average daily serving cost of 43,536 H800 GPU hours per day (1.56\% of its training cost). By improving throughput by 98\%, \NAME~reduces serving costs by 49.5\% ($1-\frac{1}{1+0.98}$), which translates to 0.77\% of the total training cost. This reduction enables the training cost for fragment tokenization method to be offset within a single day. It is important to note that Deepseek is used here as an illustrative example. Other companies may incur higher serving costs depending on their specific deployment scenarios. Nevertheless, the cost efficiency of EFIM remains a compelling advantage for scaling LLM inference.
\section{RELATED WORK}

\textbf{Cross-request KV cache reuse.} Cross-request KV cache reuse is a key feature in LLM inference framework \cite{vLLM,sglang}, aimed at reducing computation during the prefill stage. Several studies \cite{CachedAttention,PromptCache,ChunkAttention,Pensieve} have addressed the challenge of limited GPU memory for storing KV cache by utilizing CPU host memory or even disk storage to expand capacity. While these approaches focus on leveraging physical resources to improve KV cache reuse, our work improves it by transforming the prompt format in the infilling scene.

\textbf{LLMs for infilling tasks.} Using LLMs to infill contents has become a crucial technique in assisted programming, with numerous open-source models developed to support this application \cite{starcoder,starcoder2,Codellama,Qwen2.5-coder,deepseek-coder,InCoder,CodeGen,CodeGen2}. Existing research typically focuses on acquiring, curating, and generating large-scale training datasets, as well as optimizing the training process to enhance the performance and accuracy of infilling tasks.  In contrast, our work targets a specific aspect of model functionality which improves the subtoken generation ability without compromising overall model performance.

\section{CONCLUSION}

This paper identifies that the efficiency of LLM inference in infilling tasks can be hindered by the FIM format. To address this issue, we propose \NAME, a modified format designed to increase KV cache reuse. However, \NAME~reveals universal subtoken generation problems in current LLMs. To solve it, we introduce an augmented training method during data processing to empower LLMs’ sub-token generation. Experiments on two typical LLMs shows that \NAME~reduces average latency by 52\% and increases throughput by 98\%, while maintaining the model's original capabilities.

\subsubsection{Acknowledgements.} 
We are grateful to the anonymous reviewers for their helpful suggestions. Special thanks are extended to Yi Liu and Qiang Lin at Tencent for their contributions.
This research was supported by the National Natural Science Foundation of China-\#62472462/\#62402534/\#62461146204, and sponsored by CCF-Tencent Rhino-Bird Open Research Fund (CCF-Tencent \\RAGR20240102).

\subsubsection{Disclosure of Interests.}
The authors have no competing interests to declare that are relevant to the content of this article.

\bibliographystyle{splncs04}
\bibliography{ref}


\end{document}